\documentclass[sigconf]{acmart}

\usepackage{danudefs}

\AtBeginDocument{%
  \providecommand\BibTeX{{%
    \normalfont B\kern-0.5em{\scshape i\kern-0.25em b}\kern-0.8em\TeX}}}

\copyrightyear{2021}
\acmYear{2021}
\setcopyright{acmcopyright}\acmConference[SIGIR '21]{Proceedings of the 44th International ACM SIGIR Conference on Research and Development in Information Retrieval}{July 11--15, 2021}{Virtual Event, Canada}
\acmBooktitle{Proceedings of the 44th International ACM SIGIR Conference on Research and Development in Information Retrieval (SIGIR '21), July 11--15, 2021, Virtual Event, Canada}
\acmPrice{15.00}
\acmDOI{10.1145/3404835.3463027}
\acmISBN{978-1-4503-8037-9/21/07}

\acmSubmissionID{1376}

\begin{document}


\title{\textsc{Backretrieval}: An Image-Pivoted Evaluation Metric for Cross-Lingual Text Representations Without Parallel Corpora}


\author{Mikhail Fain}
\affiliation{%
  \institution{Cookpad Ltd.}
  \city{Bristol}
  \country{UK}
}
\email{mikhail-fain@cookpad.com}

\author{Niall Twomey}
\affiliation{%
  \institution{Cookpad Ltd.}
  \city{Bristol}
  \country{UK}
}
\email{niall-twomey@cookpad.com}

\author{Danushka Bollegala}
\affiliation{%
  \institution{The University of Liverpool}
  \city{Liverpool}
  \country{UK}}
\email{danushka@liverpool.ac.uk}

\renewcommand{\shortauthors}{Fain, et al.}

\begin{abstract}
Cross-lingual text representations have gained popularity lately and act as the backbone of many tasks such as unsupervised machine translation and cross-lingual information retrieval, to name a few. However, evaluation of such representations is difficult in the domains beyond standard benchmarks due to the necessity of obtaining domain-specific parallel language data across different pairs of languages. In this paper, we propose an automatic metric for evaluating the quality of cross-lingual textual representations using images as a proxy in a paired image-text evaluation dataset. Experimentally, \textsc{Backretrieval} is shown to highly correlate with ground truth metrics on annotated datasets, and our analysis shows statistically significant improvements over baselines. Our experiments conclude with a case study on a recipe dataset without parallel cross-lingual data. We illustrate how to judge cross-lingual embedding quality with \textsc{Backretrieval}, and validate the outcome with a small human study.
\end{abstract}




\fancyhead{}
\maketitle

\section{Introduction and Related Work}
\label{section:introduction}

The problem of building cross-lingual document embeddings addresses the issue of creating fixed-dimensional dense document representations which could be comparable across languages. Such methods are utilized in many downstream tasks such as unsupervised machine translation, cross-lingual information retrieval and few-shot learning for low-resource languages \cite{laser,labse}. There exist a variety of approaches for creating cross-lingual document embeddings using the existing or mined parallel language data \cite{museGoogle,xlm,labse,laser}, as well as fully unsupervised methods \cite{Bert}. Further, one can also employ monolingual document representation methods in conjunction with pretrained cross-lingual word embeddings \cite{artetxe-etal-2017-learning,lample2018unsupervised} or translation models \cite{googletranslate}. Some of the methods are more appropriate than others given a particular use case \cite{ruder2019unsupervised}, and thus having a reliable way to evaluate cross-lingual embeddings is crucial for understanding which approaches are better suited for a given domain.

The cross-lingual document representation learning approaches are typically evaluated on a standard set of benchmarks \cite{xnli,ziemski-etal-2016-united,laser,zweigenbaum-etal-2017-overview,cer-etal-2017-semeval}. Echoing the monolingual representation learning evaluation methods \cite{senteval}, the standard way to assess the quality of cross-lingual representation is to see whether documents that are close to each other in the embedding space are semantically relevant to each other \cite{cer-etal-2017-semeval,laser,labse}. Cross-lingual retrieval task provides a simple way to quantify this notion: given a query document in one language, one can rank candidate documents in another language using cosine similarity between their embeddings, and use easily interpretable ranking metrics like Recall@K \cite{museGoogle} to compare the performance between different methods.

However, this strategy as well as other approaches \cite{xnli} require the availability of a set of correspondences for documents across languages. In practice, access to bilingual annotators familiar with the domain is often required, presenting challenges even for moderately-sized datasets, especially if many different language pairs need to be evaluated. There have been attempts to mine or generate surrogate paired data by using methods such as machine translation \cite{uszkoreit-etal-2010-large}, but using such data in evaluation biases model selection towards the methods correlated with the mining strategy itself. Moreover, even if one obtains exact human translations of the documents as in \cite{laser,xnli}, the evaluation might still be biased towards machine translation methods as noted in \cite{labse}.

In this work, we propose to circumvent these problems with a novel evaluation strategy we name \textsc{Backretrieval}, applicable in the domains without parallel cross-lingual data but with some paired text-image data within languages. We note that our focus is solely on evaluation, as opposed to using images to improve the cross-lingual models as in Multimodal Machine Translation \cite{barrault2018findings}.

We use multilingual datasets with associated images to show that \textsc{Backretrieval} evaluation metric is highly correlated with standard cross-lingual retrieval metrics computed with parallel sentences in different languages, and thus is a suitable surrogate retrieval metric for such domains. We further demonstrate the application of this metric on a real-world problem domain without any parallel cross-lingual documents, namely, cross-lingual recipe retrieval. We develop a method of embedding the recipe documents across five languages which outperforms the baselines on our proposed metric. We validate this approach with a small-scale blinded human study.

\section{Method: \textsc{Backretrieval}} \label{section:method}

To formally describe our proposed evaluation metric for cross-lingual text embeddings, let us assume that we are given an \textit{evaluation} dataset of texts paired with images in the source and the target languages. Specifically, we assume the availability of a dataset $\Omega_\cS$ consisting of text-image pairs, $(T_\cS, I_\cS)$, where $T_\cS$ is a source language text and $I_\cS$ is the image associated with $T_\cS$. 
Such datasets are available in abundance, for example image captions in social media or image-annotated recipes in cooking portals \cite{Salvadora}.

Likewise, for the target language, we assume a similar dataset $\Omega_\cT$ consisting of text-image pairs $(T_\cT, I_\cT)$, where $T_\cT$ is a target language text and $I_\cT$ is the image associated with $T_\cT$.

We further assume that we are given an image similarity method $g(I_1, I_2$), which is a function that returns a similarity score of images $I_1$ and $I_2$.
Our proposed method does not assume any underlying properties of $g$. In this work, we take $g$ to be the cosine similarity between the vectors extracted from the last layer of a convolutional network pretrained on ImageNet \cite{imagenet_cvpr09} for both images \cite{RetrievalCNN}.

\begin{figure}[t]
    \centering
    \includegraphics[width=0.9\columnwidth]{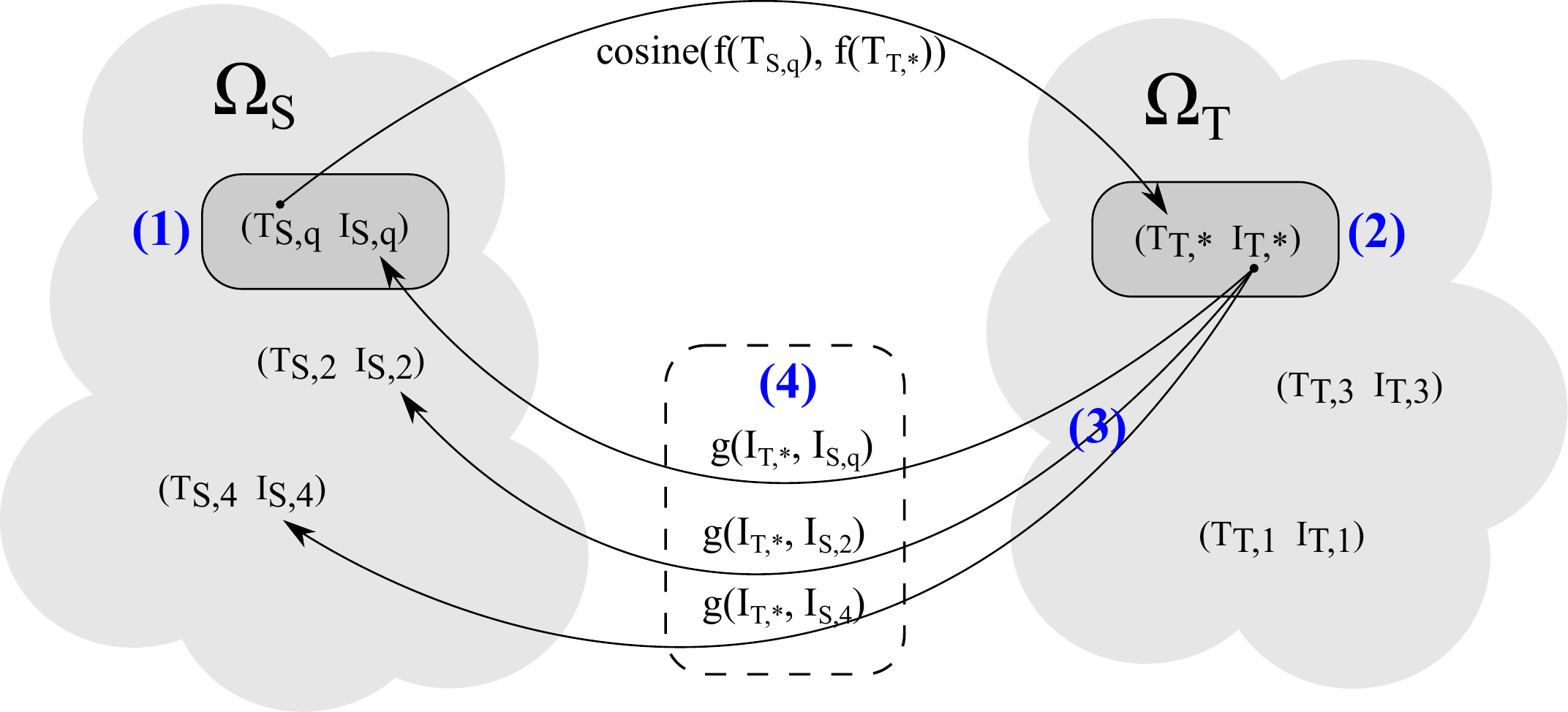}
    \caption{This image shows \textsc{Backretrieval} pictorially and explicitly annotates the four steps of the procedure.}
    \Description{The target text closest to the query source text is found via cosine similarity, then the source images are ranked by comparing them to the retrieved target image.}
    \label{fig:backretrieval-diagram}
\end{figure}

The subject under evaluation here is a cross-lingual text embedding method, $f$. Formally, $f$ is a function that takes a source or target language text $D$ as input and outputs a cross-lingual embedding vector $f(D) \in \R^n$. Our proposed evaluation metric for $f$, \textsc{Backretrieval}, is computed in four steps as follows. The procedure is also depicted in Figure \ref{fig:backretrieval-diagram}.

\begin{enumerate}
    \item For each query text-image pair $(T^q_\cS, I^q_\cS) \in \Omega_\cS$ in the source language, we first project all texts in the target language $T_\cT$ to $\R^n$ using $f$.

    \item Next, we find the most similar target language text embedding $f(T^*_\cT)$ for $f(T^q_\cS)$. We refer to $T^*_\cT$ as the \emph{retrieved text}. We denote the image corresponding to $T^*_\cT$ as $I^*_\cT$. 

    \item Next, we perform a \textsc{Backretrieval} step where we measure the similarity between $I^*_\cT$ and each $I_\cS\in \Omega_\cS$ using $g$. Then, we sort all images in $\Omega_\cS$ in the descending order of their similarity with $I^*_\cT$.

    \item Finally, we define the \textsc{Backretrieval} metric $\textsc{BkR}_f$ using the rank of $I^q_\cS$ in the sorted list. For example, we can use $Recall@K$ metric, which would report the proportion of query documents $T^q_\cS$ which had the rank of $I^q_\cS$ less than or equal to $K$. Throughout this work, we use $K=10$.
\end{enumerate}

\textsc{Backretrieval} metric is inherently an approximation to cross-lingual document retrieval metrics. Hence for this approach to be useful, the \textsc{Backretrieval} procedure should assign higher scores to the document representation models which perform well on an (unknown) ground truth cross-lingual document retrieval task, and lower scores to the document representation models which perform poorly on such a ground truth task.

We have early theoretical motivation for \textsc{Backretrieval}, and sketch our formulation here.
Assume the image-based retrieval induced by $g(\cdot, \cdot)$ performing better than random, a sufficiently large evaluation set with ground truth cross-lingual pairing, two text embedding models $f_1$ and $f_2$ and a ground truth cross-lingual retrieval metric $\textsc{XLR}$ with $\textsc{XLR}_{f_1} > \textsc{XLR}_{f_2}$. We can then show that $\textsc{BkR}_{f_1} > \textsc{BkR}_{f_2}$ with high probability if we also assume that the probability of finding the image $I^{q}_{\cS}$ during the \textsc{Backretrieval} step 4 depends only on the ground truth pairing of $I^{*}_{\cT}$. 

We note that the first assumption on $g$ is very weak and suggests that \textsc{Backretrieval} could be applicable even with relatively poor image retrieval solutions. We plan to conduct the experiments exposing the requirements on image representations quality for \textsc{Backretrieval} in future work.

Further, by analyzing the scenario where some target text embeddings are the nearest neighbours of many different source embeddings at the same time, we can show \textsc{Backretrieval} to be robust against a common issue of hubness in high-dimensional embedding spaces. Our theory is still under development, yet it nonetheless provides a theoretical foundation for \textsc{Backretrieval}.

\section{Experiments} \label{section:results}

\subsection{Datasets with Ground Truth}

In this section we evaluate a proposed evaluation metric itself rather than assessing cross-lingual models. Thus, we must find cross-lingual datasets which adhere to particular set of requirements. First, the cross-lingual datasets should have images associated to texts in different languages for the \textsc{Backretrieval} procedure to be applicable. Second, we must be able to compute the actual cross-lingual retrieval metrics to see whether our \textsc{Backretrieval} approximation is correlated with them.

We have identified three public datasets which fit these properties because they each contain a set of images with associated descriptions in different languages. \textsc{Multi30K} \cite{multi30k} extends the popular Flickr30k \cite{flickr30k} dataset with the independently generated image descriptions in German, while \textsc{MS-COCO Stair} \cite{Yoshikawa2017} extends MS-COCO Captioning dataset \cite{mscoco} with independent image descriptions in Japanese. \textsc{IKEA-Dataset} \cite{zhou-etal-2018-visual} contains IKEA furniture item photos with English, French and German descriptions.

For each dataset, the evaluation procedure is conducted as follows. We select a set of pre-trained cross-lingual text representation models of different types and expected performance. For each model $f$ we compute a pair of metrics, $(\textsc{XLR}_f, \textsc{BkR}_f)$, where $\textsc{XLR}_f$ denotes the actual cross-lingual retrieval metric computed using the matching descriptions of images in different languages, and $\textsc{BkR}_f$ denotes the \textsc{Backretrieval} metric. 

To compute $\textsc{XLR}_f$, we sample a set of $N$ documents in the source language $\cS$ along with a matching set of $N$ documents in the target language $\cT$. We embed all the documents with the model $f$, and for each source document compute a set of similarity scores with all the target documents by using cosine similarity. 
We then report the Recall@10 metric, i.e. whether the matching target document was ranked in the top 10 among all the target documents in the sample, averaged across all the source documents.

To compute $\textsc{BkR}_f$, we randomly sample a set of $N$ documents in the source language $\cS$ and a \textit{non-matching} set of $N$ documents in the target language $\cT$, simulating the scenario when we are not aware of the correspondences between documents. We then apply \textsc{Backretrieval} procedure as described in Section \ref{section:method}. We use ResNet-50 \cite{resnet50} pretrained on ImageNet \cite{imagenet_cvpr09} to induce the image similarity measure $g$ as described in Section \ref{section:method}.

For sampling, we used $N=10\,000$ for \textsc{Multi30K} and \textsc{Stair} datasets to reduce the noise in the metrics, and $N=1\,500$ for \textsc{}{IKEA-Dataset} due to its smaller size. 

After computing a set of metric pairs $(\textsc{XLR}_f, \textsc{BkR}_f)$, we treat them as data points indexed by $f$. We use Pearson and Spearman correlation coefficients for these data as a measure of a quality of approximation of $\textsc{BkR}_f$ against $\textsc{XLR}_f$. Since the procedure involves sampling, we re-run the experiments with 25 different random seeds and report the average. 

As a simple baseline to compare with \textsc{Backretrieval}, we also approximate $\textsc{XLR}_f$ by computing Spearman correlation coefficient $\textsc{CORR}_f$ between the text cosine distances and the image cosine distances, where the distances were computed between the source and target embeddings generated in the same way as for \textsc{Backretrieval}. Note that this Spearman coefficient is completely unrelated to the Spearman coefficient used for computing the quality of approximation against $\textsc{XLR}_f$.

\begin{figure}[t]
    \centering
    \includegraphics[width=0.7\columnwidth]{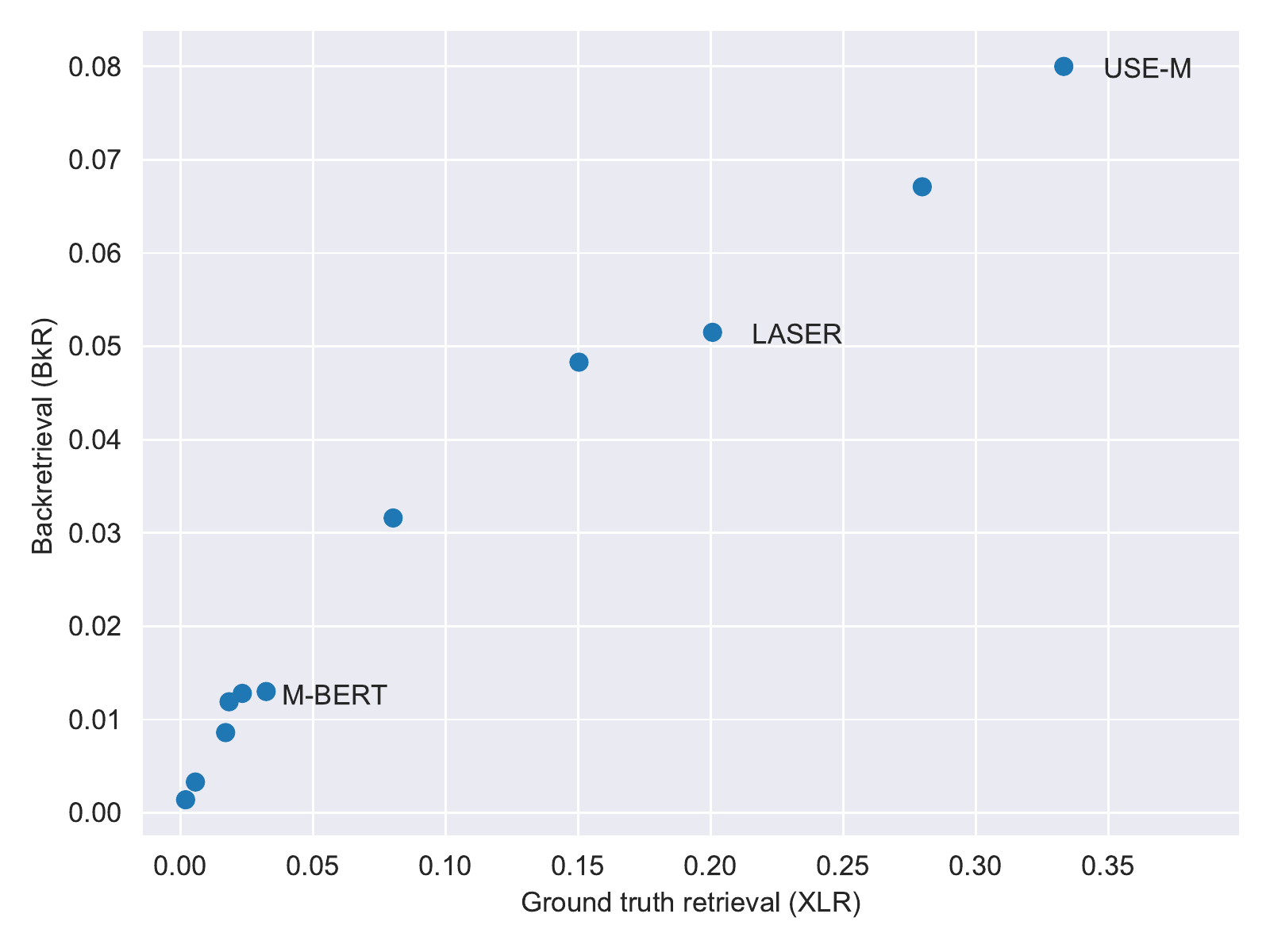}
    \caption{A scatter plot of ground truth \textit{vs.} \textsc{Backretrieval} metrics from a variety of models on the \textsc{Stair} dataset ($\text{en}\rightarrow\text{jp}$; M-BERT, LASER and USE-M are highlighted). Two correlation measures are used to assess how well \textsc{Backretrieval} estimates ground truth: Pearson (measuring linear dependence) and Spearman (measuring similarity of rank-orderings). Complete results are shown in Table \ref{table:results41}.}
    \Description{The scatter plot resembles a line, demonstrating that increased performance along the x-axis (ground truth) correlates well with the proposed approximation metric (Backretrieval).}
    \label{fig:correlation}
\end{figure}

The following cross-lingual representation methods and their variations have been used as data points $f$ in our evaluation: RCSLS \cite{joulin-etal-2018-loss}, M-BERT \cite{Bert}, mDistilBERT \cite{Sanh2019DistilBERTAD}, LaBSE \cite{labse}, LASER \cite{laser}, USE-M \cite{museGoogle}, MUSE \cite{conneau2017word}, GloVe \cite{pennington-etal-2014-glove} with translations \cite{anastasopoulos19embeddings,googletranslate}, XLM-MLM \cite{xlm}, XML-Roberta \cite{xlmRoberta}. We report the average performance of \textsc{Backretrieval} and \textsc{CORR} across all datasets in Table \ref{table:results41}, and plot $(\textsc{XLR}_f, \textsc{BkR}_f)$ datapoints for one random seed in Figure \ref{fig:correlation} for illustration purposes. The data suggest that \textsc{Backretrieval} approximates the ground truth cross-lingual retrieval metrics very well on these datasets.

We emphasise that the question of which of the above methods $f$ performed better or worse for a particular dataset is irrelevant for this experiment. What matters is whether the performance $\textsc{XLR}_f$ is correlated with $\textsc{BkR}_f$.

\begin{table}[t]
\caption{The correlation of the proposed method (\textsc{BkR}) and the baseline (\textsc{CORR}) with the ground truth metric (\textsc{XLR}), across datasets. The best results are in bold. The difference is statistically significant across all the reported metrics.}
\label{table:results41}
\begin{center}
\setlength{\tabcolsep}{0.5 em}
\begin{tabular}{l rrrr}  
\toprule
Dataset & \multicolumn{2}{c}{Pearson}  & \multicolumn{2}{c}{Spearman}  \\
\midrule
 & \textsc{BkR} & \textsc{CORR} & \textsc{BkR} & \textsc{CORR} \\
\midrule
\textsc{Stair} en$\rightarrow$jp & \textbf{.98$\pm$.001} & .76$\pm$.002 & \textbf{.98$\pm$.003} & .91$\pm$.003 \\
\textsc{Stair} jp$\rightarrow$en & \textbf{.98$\pm$.001} & .75$\pm$.001 & \textbf{.97$\pm$.003} & .90$\pm$.005 \\
\textsc{Multi30k} en$\rightarrow$de & \textbf{.99$\pm$.001} & .81$\pm$.003 & \textbf{.98$\pm$.002} & .93$\pm$.002 \\
\textsc{Multi30k} de$\rightarrow$en & \textbf{.99$\pm$.001} & .79$\pm$.002 & \textbf{.97$\pm$.002} & .85$\pm$.004 \\
\textsc{IKEA} en$\rightarrow$de & \textbf{.97$\pm$.001} & .69$\pm$.007 & \textbf{.98$\pm$.002} & .63$\pm$.009 \\
\textsc{IKEA} de$\rightarrow$en & \textbf{.97$\pm$.001} & .73$\pm$.007 & \textbf{.95$\pm$.003} & .67$\pm$.010 \\
\textsc{IKEA} en$\rightarrow$fr & \textbf{.98$\pm$.001} & .69$\pm$.007 & \textbf{.96$\pm$.002} & .68$\pm$.007 \\
\textsc{IKEA} fr$\rightarrow$en & \textbf{.97$\pm$.001} & .76$\pm$.006 & \textbf{.92$\pm$.004} & .71$\pm$.008 \\
\bottomrule
\end{tabular}
\end{center}
\end{table}

\subsection{Real-world Case Study with No Ground Truth}

In this section, we illustrate how to make use of \textsc{Backretrieval} evaluation for developing cross-lingual representations when no cross-lingual correspondences are available. We emphasise that the purpose of this section is to demonstrate the use of \textsc{Backretrieval} in the appropriate context rather than developing novel methods. We address the problem of \textit{cross-lingual recipe representation}, where the goal is to represent a cooking recipe in a language-independent way. As an example, a \textit{`chicken soup'} recipe in English should have the representation which is close to \textit{`sopa de pollo'} Spanish recipe in the embedding space. In most cases, however, the definition of similar cooking recipes is not clear cut.

For evaluating the cross-lingual representations, we use a dataset of cooking recipes along with their corresponding images in five languages Arabic (ar), English (en), Indonesian (id), Russian (ru), Spanish (es), retrieved from the online cooking platform Cookpad. For each language, we extracted $10\,000$ unique recipes. These are not parallel corpora, so correspondence information is unavailable.

We first evaluate a set of pre-trained cross-lingual models with \textsc{Backretrieval}. For 5 languages, we have 20 different induced cross-lingual retrieval tasks, and we compute the \textsc{Backretrieval} metric for each of the tasks for each method with  ResNext-101 \cite{Xie2016} last convolutional layer as the image representation. To make the discussion compact,
we average the performance across different language pairs per method and report the results in Table \ref{table:results42}. Based on these results, we choose LaBSE \cite{labse} as the model to build upon.

\begin{table}[t]
\caption{\textsc{Backretrieval} performance on Cookpad dataset for different models, averaged across all language pairs. The best results are in bold. The difference between the best models and others is statistically significant.}
\label{table:results42}
\begin{center}
\begin{tabular}{l r}  
\toprule
Method & \textsc{BkR} ($\times 1000$)  \\
\midrule
Random & 1.0 \\
mDistilBERT \cite{Sanh2019DistilBERTAD} & 1.2 \\
RCSLS \cite{joulin-etal-2018-loss} & 1.7 \\
LASER \cite{laser} & 2.8 \\
GloVe+Translation \cite{pennington-etal-2014-glove,anastasopoulos19embeddings} & 3.4 \\
LaBSE \cite{labse} & 5.2 \\
\midrule
\textbf{\textsc{Triplet} (Ours)} & \textbf{9.3} \\
\textbf{\textsc{TripletXL} (Ours)} & \textbf{9.4} \\
\bottomrule
\end{tabular}
\end{center}
\end{table}

To improve upon the baseline LaBSE's performance, we have sourced additional $100\,000$ textual recipes (without images) for each of the five languages from the same platform, and trained a model denoted \textsc{Triplet} on this dataset on top of LaBSE representations. In particular, we first embedded the recipe title and recipe body (ingredients and instructions) separately with LaBSE. We then used a siamese architecture with two dense layers with ReLU activations applied to both title and the body, 
resulting in two 300-dimensional representations for earch recipe $r$, one for the title ($x_{rt}$) and one for the body ($x_{rb}$). We trained this model with triplet loss: 

\begin{equation}
\label{eq:triplet}
\cL_{\textsc{Triplet}}^r = \max(0, d(x_{rb}, x_{rt}) - d(x_{rb}, x_{nt}) + \alpha)
\end{equation}

\noindent where $x_{nt}$ is a random recipe title representation from the mini-batch used as a negative,  $d$ is the cosine distance between two vectors and $\alpha$ is the margin.

The model was trained for 40 epochs with Adam optimizer \cite{kingma2014adam}, learning rate 0.01, $\alpha=0.1$ and batch size 128. The resulting representation used in the evaluation for a recipe $r$ is the concatenation of $x_{rt}$ and $x_{rb}$. We have used a random 90-10 train-validation split for hyperparameter tuning.

\begin{figure}[t]
    \centering
    \includegraphics[width=0.9\columnwidth]{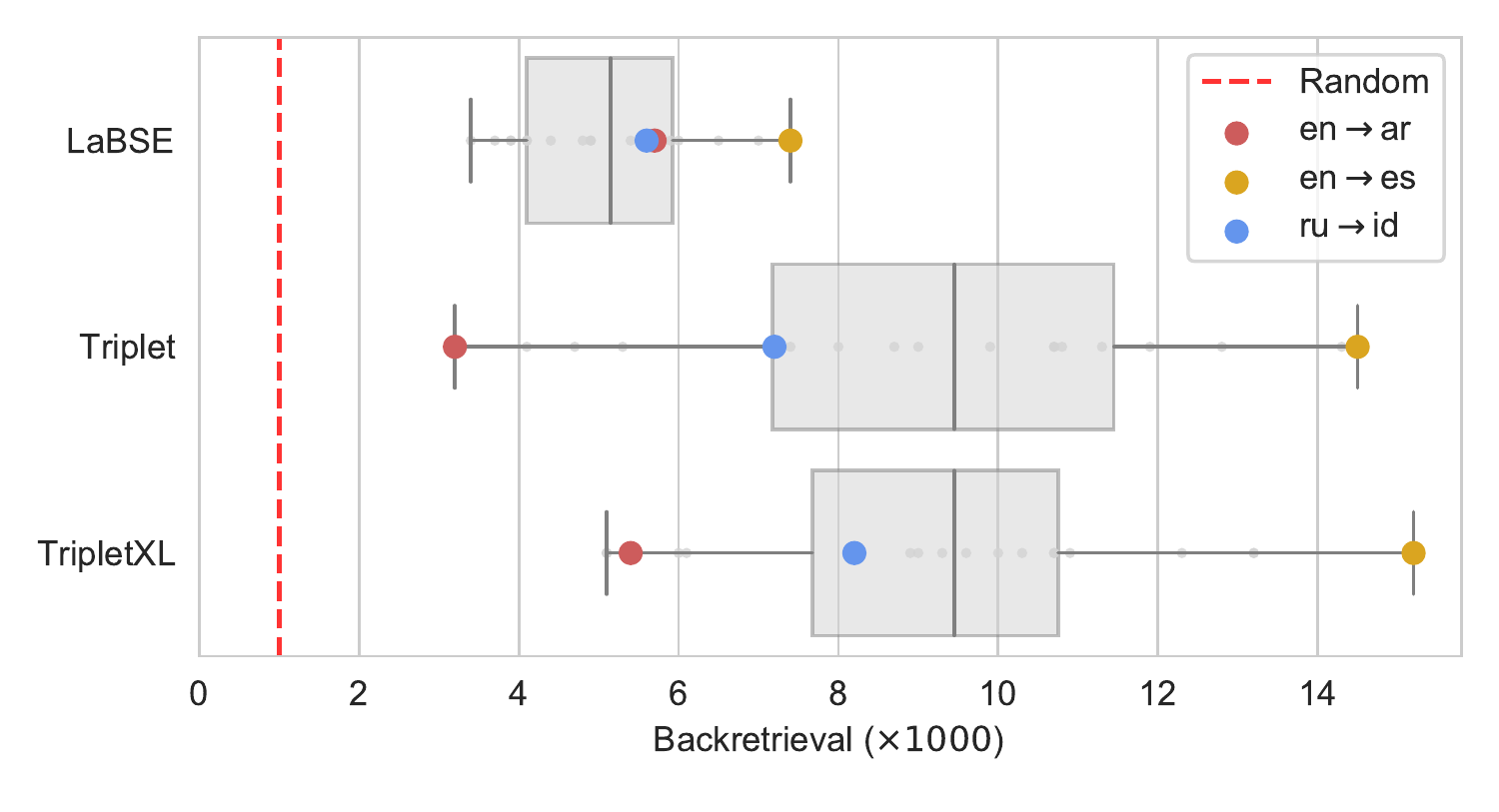}
    \caption{The boxplot of \textsc{Backretrieval} performance on Cookpad dataset for different models. Each data point is a language pair for a particular model, with en$\rightarrow$ar, en$\rightarrow$es and ru$\rightarrow$id annotated. While \textsc{Triplet} improves upon \textsc{LaBSE} on average, the \textsc{Triplet} spread of \textsc{Backretrieval} metric across language pairs is high. \textsc{TripletXL} modification reduces the spread and improves the performance on the most challenging language pairs for this model (e.g., en$\rightarrow$ar)}
    \Description{The boxplot mean is higher for Triplet and TripletXL than for LaBSE. The boxplot shows lower variance for TripletXL vs Triplet.}
    \label{fig:boxplot}
\end{figure}

We report the average \textsc{Backretrieval} performance across language pairs in Table \ref{table:results42}, with Figure \ref{fig:boxplot} showing more details of this distribution. The results show that although \textsc{Triplet} average improves greatly over LaBSE, the new model performs much worse for some language pairs. We hypothesize that this is due to $\cL_{\textsc{triplet}}$ loss not enforcing any cross-lingual constraints, which could lead to one language being mapped to a different part of the resulting embedding space than other languages. We thus augment our loss function with an additional term, which penalizes the representations that are much closer to their neighbours in the same language than to their neighbours in other languages:

\begin{equation}
\label{eq:langaw}
\begin{split}
& \cL_{\textsc{TripletXL}}^r = \cL_{\textsc{Triplet}}^r + \beta \cL_{\textsc{XL}}^r \\ 
& \cL_{\textsc{XL}}^r = \textstyle \sum_{l \neq l_r}{\left| \frac{1}{K}\sum_{n \in \cN_l}d(x_{rb},x_{nt}) - \frac{1}{K}\sum_{n \in \cN_{l_r}}d(x_{rb},x_{nt}) \right|}
\end{split}
\end{equation}

\noindent where $l_r$ is recipe $r$'s language, $\cN_l$ is the set of $K$ nearest neighbours recipes of language $l$ for the recipe $r$, and $\beta$ is a hyperparameter.

We have trained our model with $\beta=0.01$ and $K=5$, with the rest of the training setup identical to the training of the \textsc{Triplet} model, and refer to this new setup as \textsc{TripletXL} in Table \ref{table:results42} and Figure \ref{fig:boxplot}. We observe that \textsc{TripletXL} average performance remains close to \textsc{Triplet}, but the problem of extremely poor performance for some language pairs has been mitigated.

To validate the performance of \textsc{TripletXL} model beyond \textsc{Backretrieval}, we have conducted a small-scale blind randomized trial with one volunteer.
For 20 different English recipes from the evaluation set, the subject was shown 10 suggested Russian recipes in random order: 5 extracted from the test set using LaBSE model, and 5 extracted using \textsc{TripletXL}. The subject scored each suggestion on a scale from 1 to 4 (``very relevant" to ``not relevant""). The results showed that \textsc{TripletXL} extracted 2.3 times more ``very relevant" suggestions on average, in line with \textsc{Backretrieval} relative performance improvement. The results are statistically significant. We re-iterate that the purpose of this case study was not to thoroughly assess the novel method, but rather to demonstrate \textsc{Backretrieval}'s utility
for evaluating models contextually.

\section{Conclusions} \label{section:conclusions}

Early analysis for a simple yet efficient novel cross-lingual evaluation metric called \textsc{Backretrieval} is described in this paper. Our key contribution is to show that within-language text-image pairs can be successfully leveraged for evaluating cross-lingual representations in the absence of parallel corpora. Our experimental analysis is strongly suggestive of the utility of our approach: the proposed \textsc{Backretrieval} metric correlates very highly with the ground truth on parallel corpora across three public datasets. The case study experiments on the recipe retrieval task without parallel data are also validated by a blinded human study. By introducing an evaluation metric that abandons requirements for parallel corpora, we hope to expand and accelerate research on cross-lingual document embeddings. Future work will formulate theoretical justifications of the approach more rigorously and increase the experimental validation, specifically exploring whether \textsc{Backretrieval} scores can be calibrated to align with ground truth metrics.

\bibliographystyle{ACM-Reference-Format}
\bibliography{refs}


\end{document}